%% file: paper.tex
\documentclass{article}
\usepackage[T1]{fontenc}
\usepackage{microtype}
\usepackage{amsmath,amssymb,booktabs,array}
\usepackage{graphicx}
\usepackage{framed}
\usepackage[dvipsnames]{xcolor}
\usepackage{tikz}
\usetikzlibrary{arrows.meta,positioning,fit,plotmarks}
\PassOptionsToPackage{hyphens}{url}
\usepackage[hidelinks]{hyperref}

\PassOptionsToPackage{compatV3}{fancyhdr}
\usepackage[accepted]{mlsys2025}
\usepackage{balance}
\makeatletter
\renewcommand{\Notice@String}{}
\renewcommand{\mlsysaffiliation}[2]{%
  \textsuperscript{\csname the@affil#1\endcsname}#2%
}
\makeatother
\mlsystitlerunning{Metal Sandbox: Isolated Execution and Command Scheduling for GPU Coding Agents on Apple Silicon}
\newcommand{\system}{Metal Sandbox}
\definecolor{staticcolor}{HTML}{64748B}
\definecolor{transfercolor}{HTML}{2563EB}
\definecolor{payloadcolor}{HTML}{D97706}
\newcommand{\code}[1]{\texttt{#1}}

\begin{document}
\twocolumn[
\mlsystitle{Metal Sandbox: Isolated Execution and Command Scheduling for GPU Coding Agents on Apple Silicon}
\begin{mlsysauthorlist}
\mlsysauthor{Haozhe Wang}{sjtu,zgc,stepfun}\quad
\mlsysauthor{Lina Chen}{stepfun}\quad
\mlsysauthor{Fanqi Wan}{stepfun}\quad
\mlsysauthor{Qi Han}{stepfun}\\
\mlsysauthor{Yinmin Zhang}{stepfun}\quad
\mlsysauthor{Zheng Ge}{stepfun}\quad
\mlsysauthor{Daxin Jiang}{stepfun}\quad
\mlsysauthor{Junchi Yan}{sjtu}
\end{mlsysauthorlist}
\begin{center}
\small
\mlsysaffiliation{sjtu}{Shanghai Jiaotong University}\quad
\mlsysaffiliation{zgc}{Beijing Zhongguancun Academy}\quad
\mlsysaffiliation{stepfun}{StepFun}
\end{center}
\mlsyskeywords{GPU coding agents, distributed execution, command scheduling, Apple Silicon, Metal}
\vskip 0.3in
\begin{abstract}
On-device inference requires kernels that make effective use of local accelerators.
Efficient training and evaluation of models that generate Apple Metal kernels benefit from native macOS tools and concurrent agent sessions.
These sessions alternate between model requests, development commands, and GPU execution, leaving GPU capacity idle for much of a rollout.
We present \textbf{\system{}}, infrastructure that combines restricted per-instance file access, native command execution, and command-level scheduling across interconnected Macs.
Each agent retains its workspace and private development environment.
Parallel commands can run concurrently, while serial commands share one execution slot per node; eligible waiting commands can migrate to available Macs and return results to the original agent.
Integration with Harbor supports task execution, automatic verification, and rollout collection through a common workflow.
Controlled experiments on four Mac mini M4 nodes achieve 1.68$\times$--2.05$\times$ gains in command throughput under concentrated demand.
Sixteen static agent traces give an estimated 2.14$\times$ sustained-throughput gain for four sessions per node versus one, assuming continuously replenished sessions and unchanged mean durations.
Replaying these measured traces under dynamic scheduling reduces mean tool waiting by 18.8\% and batch elapsed time by 22.3\%, assuming unchanged phase durations.
Metal Sandbox and its Harbor integration are open-source on GitHub.{\renewcommand{\thefootnote}{*}\footnotemark}
\end{abstract}
]
\begingroup
\renewcommand{\thefootnote}{*}
\footnotetext{\begin{tabular}[t]{@{}l@{}}
\href{https://github.com/WangHaoZhe/metal-sandbox}{\nolinkurl{github.com/WangHaoZhe/metal-sandbox}}\\
\href{https://github.com/WangHaoZhe/harbor}{\nolinkurl{github.com/WangHaoZhe/harbor}}
\end{tabular}}
\endgroup

\section{Introduction}
Efficient on-device inference requires kernels that use local accelerators well. Apple's on-device foundation model illustrates this need for local inference optimization~\cite{applefoundation}. Metal provides a native GPU API and shading language for both M-series chips in Macs and A-series chips in iPhones~\cite{metaloverview,metalfeatures}. These processors combine high compute performance with energy-efficient designs~\cite{applea18,applem4}; unified memory also supports CPU/GPU workflows in frameworks such as MLX~\cite{mlx}. This combination makes Metal important for on-device computation. Improving models' ability to generate efficient Metal kernels can support applications across Apple's desktop and mobile devices.

Language models can generate and iteratively improve GPU kernels using compiler, correctness, and performance feedback. KernelBench provides evaluation tasks for CUDA and GPU-DSL generation~\cite{kernelbench}. KernelGYM, introduced with \mbox{Dr.\ Kernel}, supplies distributed GPU evaluation and multi-turn data collection for reinforcement learning~\cite{drkernel}, while Metal-Sci studies automatic search over native Metal scientific kernels~\cite{metalsci}. Supporting broader Metal evaluation and training requires interactive agents to prepare dependencies, edit programs, compile candidates, and obtain feedback across concurrent rollouts.

These agents alternate between model requests, source editing, CPU preparation, and candidate execution. Figure~\ref{fig:single_agent_motivation} measures one \code{metal\_nbody} session (task specification in Appendix~\ref{app:nbody_task}) without competing agents. Apple Instruments' Metal System Trace~\cite{metaltools} records 44.0 seconds of GPU activity over 31.6 minutes, or 2.32\% of the session. Most of the session is spent outside GPU execution. We therefore schedule individual commands across concurrent sessions to use idle GPU capacity, reduce tool waiting, and shorten rollout collection. Commands that run or time GPU work enter a serial queue on each Mac, while parallel commands, such as file inspection and editing, proceed without waiting for that queue. This allows commands to overlap while coordinating participating GPU timing commands. Per-node serialization alone, however, leaves queues unbalanced across Macs. In sixteen concurrent sessions on four Macs (Section~\ref{sec:agent_rollout}), tool waiting accounts for 21.0 of the 45.1 minutes in an average session. Although each Mac hosts four sessions, sessions on the busiest Mac wait 32.7 minutes on average, compared with 13.3 minutes on the least busy Mac. Waiting serial commands can therefore migrate to Macs with free execution slots.

\input{artifacts/figure_01/figure.tex}

Concurrent development also requires separate instance files and a usable native tool environment. Agents need shell commands, compilers, and Metal programs without unrestricted write access to the host user's files. Table~\ref{tab:execution_choices} compares these environments. Docker containers and macOS VMs provide separate filesystems; bare macOS has no built-in file isolation between agent sessions. Standard Docker Desktop containers execute inside a Linux VM~\cite{dockervmm}, and Docker Desktop's GPU support requires Windows with the WSL~2 backend~\cite{dockergpu}. macOS VMs provide a guest environment and virtual Metal graphics~\cite{macvirtualization}. Bare macOS provides native execution, with file restrictions and timing coordination left to the application. KernelGYM already serializes evaluation tasks per GPU, but its evaluation backend targets CUDA~\cite{drkernel}. Interactive Metal rollouts additionally need persistent native workspaces and scheduling of individual tool commands.

\input{artifacts/table_01/table.tex}

We present \textbf{\system{}}, native execution infrastructure that combines restricted instance file access with command scheduling across interconnected Macs. Each agent receives an instance, a workspace, and a private development \code{HOME}. Commands run in Apple's App Sandbox, an operating-system mechanism that restricts an application's access to files and other resources~\cite{appsandbox}, with access to the instance's files and required system tools. Our Harbor integration supports Claude Code, Codex, and mini-SWE-agent, enabling configured tasks, automatic verification, and trajectory collection through the same evaluation workflow~\cite{harbor,minisweagent}.

The paper contributes:
\begin{enumerate}
\item A native multi-agent execution model with restricted per-instance file access, ordinary development commands, and per-node serial queues, integrated with Harbor for task execution and rollout collection.
\item Cross-Mac command migration that shares instance development files, coordinates execution at source and target nodes, and returns results to the original agent.
\item A four-Mac evaluation of cluster scaling, concentrated demand, and migration cost, plus scheduling replay that holds sixteen measured agent trajectories fixed.
\end{enumerate}

\section{Native Execution and Instance Isolation}
\label{sec:model}
\subsection{Native Command Interface}
An \emph{instance} provides a workspace and development environment for an agent session, which repeatedly submits tool commands. Its \emph{owner node} is the Mac hosting that session; an \emph{execution node} is the Mac currently running one of its commands. During migration, the source node holds the waiting command, and the target node receives it. Each managed command includes the submitted program or shell expression and its child processes. The management service executes candidate builds and runs directly on host macOS in an App Sandbox process; ordinary child processes inherit its sandbox policy. Agents use native shell commands, installed compilers, SDKs, and Metal through this interface.

\subsection{Instance Filesystem Isolation}
The runtime maps \code{HOME} and shell \code{\textasciitilde} to the instance's development directory and starts commands in its workspace. Agents can create and modify files in that workspace, private \code{HOME}, and temporary directories. Required system tools, SDKs, and configured development-tool directories remain accessible. Each sandbox process receives access to its own instance's development files, without access to the host user's entire home directory or other instances' files.

The following commands were executed in one instance using native \code{/bin/sh} and \code{/bin/mkdir}. The first creates a directory in the instance's \code{HOME}; the second attempts the same operation in the host user's home.

\input{artifacts/native_command_example/example.tex}

This per-instance filesystem isolation lets an agent edit, build, and test its own project while rejecting writes to host files outside the granted scope and to other instances' workspaces.

\subsection{Shared Files for Command Migration}
\label{sec:shared_files}
For dynamic scheduling, each instance uses its own shared workspace and development \code{HOME}, mounted at the same absolute paths on nodes in its scheduling pool. This makes the instance's files available wherever its commands execute, while different instances retain separate directories. Temporary files, runtime binaries, and execution logs remain local. Identity markers detect shared directories that are missing, replaced, or mounted incorrectly. Toolchains, dependencies, and permissions must be compatible across nodes.

Migration starts a new process on the target node for a waiting command; it does not move a running process, live Metal buffers, or GPU state between Macs. Shared storage does not coordinate simultaneous writes by different commands. Commands that require local GUI state, sockets, running processes, unshared files, or device-specific state remain on their owner node.

\section{System Architecture}
\label{sec:architecture}
Figure~\ref{fig:architecture} shows how commands, results, and shared files move through the system. Figure~\ref{fig:cluster_deployment} shows the four-Mac deployment, with Harbor and the cluster controller on the management node. The controller accepts cluster requests and records instance and command locations. Each node runs its own scheduler and sandbox processes, and checks saved execution state at startup to identify unfinished work left by an unexpected shutdown before accepting new work. Migrated commands execute through the target node's normal management service.

\input{artifacts/figure_02/figure.tex}

\input{artifacts/figure_03/figure.tex}

\subsection{Cross-Node Communication and Location Tracking}
When a request concerns another Mac (for example, placing an instance on a worker or migrating a waiting command), the cluster controller connects over SSH and invokes that node's Metal Sandbox CLI. The remote service handles instance operations, file transfer, and execution-status queries; candidate commands run in its sandbox processes.

The controller records which node owns each instance and which node executes each command, together with migration progress. These records are saved to disk so later requests can find the same instance or command. For example, after a command migrates from Mac mini \#B to \#D, requests for its output are sent to \#D while the agent remains on \#B.

\subsection{Instance Placement}
For node $n$, let $A_n$ be the observed number of active instances and $R_n$ the number of instances assigned there but still being created. Automatic placement selects from the eligible node set $\mathcal{E}$:
\begin{equation}
n^* = \arg\min_{n\in\mathcal{E}} (A_n+R_n).
\end{equation}
An eligible node must have a compatible service version, available instance capacity, no pending recovery, and a thermal state below critical. During batch creation, the controller updates the projected counts after each assignment so successive instances do not all use the same initial counts.

Instance count balances concurrent sessions but does not predict how much GPU work each session will submit. Command migration reduces queue imbalance that remains after instance placement.

\subsection{Per-Node Command Scheduling}
The default allowlist includes common file and text commands, compilers, and build tools; Appendix~\ref{app:allowlist} gives the complete list. Matching commands run in parallel; other commands enter the serial queue. The administrator can edit this list or use a blocklist, where matching commands run serially and others run in parallel. Rules match executable paths and argument prefixes, and supported shell expressions are inspected before scheduling. Rules cover a command and its child processes; they do not detect GPU calls. A build tool may launch GPU tests, so the administrator must classify that build command as serial when those tests require coordinated timing.

Each node has one \emph{serial execution slot} and a first-in, first-out (FIFO) queue. Parallel commands can run while this slot is occupied. Queues are kept per node rather than cluster-wide. Each node admits its own serial commands without a round trip to the controller, so local scheduling adds no network latency and does not depend on controller availability. Cross-node balance is instead provided by migrating waiting commands (Section~\ref{sec:transfer}), at the cost of not enforcing a global arrival order.
Queue waiting is recorded separately and is excluded from Metal Sandbox's managed command execution budget and Harbor's active agent execution budget. The managed command's execution budget starts when it receives execution permission.

\subsection{Eligibility and Target-Node Selection}
\label{sec:transfer}
Dynamic scheduling considers only waiting serial commands from instances that enable it and meet the shared-file requirements in Section~\ref{sec:shared_files}. The command's arguments and environment must pass the source-private-path check; the target node also validates its files, environment, and scheduling rules. Parallel commands and agent sessions remain on their owner nodes. Intercepted shell tool calls supply their script, working directory, environment, relative timeout duration, and output-size limit. Calls missing this information remain local.

The controller checks node status and looks for serial commands waiting at source nodes. A target node must have a free serial slot, no local serial waiters, no pending recovery, and a thermal state below critical. Among eligible remote nodes, the controller chooses the smallest active-instance count; equal counts are resolved by selecting the smallest node ID. Reservation then holds the target node's slot so another command cannot take it before migration completes.

\subsection{Scheduling Pools}
The administrator assigns each node to a \emph{scheduling pool}, for example to group Macs with compatible hardware and toolchains. Automatic instance placement can be restricted to a selected pool. Command migration stays within the source node's pool and requires cross-node scheduling to be enabled on both source and target nodes.

\subsection{Reservation, Claim, and Commit}
Figure~\ref{fig:protocol} shows how a waiting command moves from its source queue to execution on a target node through reservation, claim, and commit. Reservation checks the shared workspace, command, environment, and scheduling rules, then holds the target node's serial slot and prepares its sandbox process. The controller saves the command ID, source node, target node, and migration progress. Claim transfers execution responsibility from the source queue to the target reservation, disabling a local start. Commit authorizes the reserved command to start on the target node.

If the source command has already started or been cancelled, claim fails and the controller releases the unused target reservation. If the source node accepted the claim but its response was lost, the controller queries the same command ID before proceeding. It then commits the reserved command to the target node, which checks the command description and reservation ID. The target node records the command as queued, then running when execution starts.

\input{artifacts/figure_04/figure.tex}

\subsection{Results and Accounting}
The target node returns stdout, stderr, exit status, and cleanup information using the migrated command's ID. The source node uses these results to finish the original CLI execution or shell tool call. Execution logs and local temporary files stay on the target node; workspace modifications are visible through shared storage.

The command remains attributed to its owner instance, while per-node execution counts record where it actually runs. Batch elapsed time is measured at the submitting controller. The controlled experiments report queue waiting at both nodes; the agent traces report the broader submission-to-start interval defined in Section~\ref{sec:agent_rollout}.

\section{Harbor Integration}
\label{sec:harbor}
Harbor connects task specifications, coding agents, and automatic verification through the Metal Sandbox environment adapter. For each trial, the adapter creates an instance, uploads the task files, prepares the agent runtime within the instance, and starts the configured harness. The agent and verifier use separate execution phases; the verifier receives the completed workspace and returns the task result. Harbor retains these results and the agent trajectory for evaluation and rollout collection.

Claude Code, Codex, and mini-SWE-agent are supported through the same managed command interface. Claude Code and Codex use tool hooks to route shell commands through the sandbox command helper. mini-SWE-agent uses its \code{MetalEnvironment} interface to submit commands and receive their outputs. Agent sessions run concurrently, and each tool call is scheduled independently. A migrated call retains its shell script, working directory, environment, and original result interface. Calls that lack a shared-workspace description remain on the owner node.

Harbor task configuration selects the harness and runtime version, model endpoint, concurrency, and execution budgets. The adapter records command waiting separately from active execution, preserves stdout, stderr, and exit status, and collects the native trajectory and standardized ATIF record. This allows rollout timing to distinguish model requests, tool execution, and tool waiting.

\section{Evaluation}
\label{sec:evaluation}
We measure command throughput, scaling with node count, and queue waiting under balanced and concentrated submissions, then examine longer commands and migration overhead. The controlled experiments use fixed Metal implementations. Section~\ref{sec:agent_rollout} reports all sixteen measured static agent trajectories; Section~\ref{sec:replay} replays those fixed traces under dynamic scheduling.

\subsection{Experimental Setup and Workloads}
We use fixed workloads to measure how command scheduling changes throughput and waiting as demand and node count vary. The experiments use four Mac mini nodes with Apple M4 processors, 32 GiB memory per node, and macOS 27.0.1 (build 26A434), interconnected through a gigabit Ethernet switch. Two owner instances are prepared per node. Shared workspaces use SMB mounts at matching absolute paths, and control requests use the remote CLI over SSH.

Table~\ref{tab:workloads} lists the six hand-written Metal kernels. Inputs are deterministic float32 arrays, with $a_i=((i\bmod23)-11)/16$ and $b_i=((i\bmod17)-8)/16$. The matrix and convolution workloads use these values in their respective shapes.

\input{artifacts/table_03/table.tex}

An Objective-C++ runner is compiled with \code{-O2} before the measured batches. Each command loads and compiles the Metal source, creates its pipeline and buffers, and repeatedly submits the selected kernel. Every submission is followed by \code{waitUntilCompleted()}, which waits for the GPU command buffer to finish before its GPU start and end timestamps are read. Five submissions warm up execution, followed by at least thirty valid GPU timestamp samples. The loop repeats submissions until the elapsed time from its start reaches the requested minimum of two or six seconds, including encoding, submission, and waiting. We call these workloads two-second and six-second commands, respectively. Both use the same small input shapes.

The runner writes each result for an external correctness check. Command outputs are checked before they contribute to the reported throughput.

\subsection{Baselines and Comparison Design}
The comparison holds the controller build, Metal implementations, participating machines, workload commands, runner, shared storage, scheduling rules, and numerical checker constant. We compare:
\begin{enumerate}
\item \textbf{Static scheduling:} commands execute on their assigned owner nodes, with cross-node command migration disabled.
\item \textbf{Dynamic scheduling:} eligible waiting serial commands can migrate to available target nodes through the reservation--claim--commit protocol.
\end{enumerate}
Owner assignments are controlled explicitly. Balanced batches distribute commands round-robin across the selected nodes; concentrated batches submit all commands through two instances on the management node. All workload commands use the serial queue, allowing one workload command per node at a time.

Batch time is measured on the submitting controller from submission through completion of all result reads and external numerical checks. Runner compilation and reference-output preparation precede that interval; per-command Metal compilation and pipeline creation occur inside it. Throughput is the completed command count divided by batch time, expressed as commands per hour. Configuration order is randomized within each repetition. We report medians and observed minimum--maximum ranges over three independent batches per configuration.

\subsection{Throughput, Scaling, and Queue Delay}
We first measure throughput, scaling, and queue delay using twelve commands per batch, with each of the six kernels executed twice and a two-second minimum computation loop. Balanced submissions use one, two, or four nodes; concentrated submissions use the four-node cluster.

Table~\ref{tab:throughput} reports batch times and throughput. Balanced static execution decreases from 25.2 seconds on one node to 6.62 seconds on four nodes. With dynamic scheduling enabled, balanced runs take 25.1, 12.8, and 6.61 seconds on one, two, and four nodes. No commands migrate in these balanced batches, so the improvement comes from parallel execution across the available nodes.

\input{artifacts/table_04/table.tex}

\input{artifacts/figure_05/figure.tex}

With concentrated submissions on the four-node cluster, median batch time decreases from 25.3 to 15.1 seconds, a 40.4\% reduction and 1.68$\times$ throughput gain. The controller initiates at most one migration per source node per scan and prepares remote execution through synchronous control requests, while the source queue continues executing locally. This limits how quickly the three remote serial slots can be filled for approximately two-second commands. Recorded GPU device identities show per-node command counts of 7/3/1/1, 7/3/1/1, and 6/3/2/1 across the three dynamic batches. Six or seven commands therefore still execute serially on the source node, compared with three per node in the balanced case. At approximately 2.10 seconds per command, this remaining source-node work accounts for approximately 12.6--14.7 seconds of execution. The remaining concentration, together with remote preparation and control costs, limits the speedup below balanced four-node scaling.

The runtime's \code{queueWaitSeconds} adds queue waiting at the source and target nodes. We compute the linearly interpolated 95th percentile within each twelve-command batch. Under concentrated submissions, the median of these batch values decreases from 21.9 [21.8, 21.9] seconds to 11.9 [10.5, 12.9] seconds, a 45.5\% reduction (Figure~\ref{fig:throughput}b).

\subsection{Long-Command Scheduling}
To examine longer commands, we increase the minimum computation-loop duration to six seconds and submit each of the six kernels once per batch, using the same submission setup as the concentrated two-second batches.

Static execution takes 36.8 [36.7, 36.8] seconds per batch, while dynamic scheduling takes 18.0 [17.6, 18.2] seconds (Figure~\ref{fig:throughput}a). This gives a 51.1\% reduction in median batch time and 2.05$\times$ throughput: 588 versus 1,200 commands per hour. Four of the six commands execute on other nodes in every dynamic batch.

Longer commands benefit from migration despite the cost of preparing remote execution.

\subsection{Migration Overhead}
\label{sec:migration_overhead}
We perform 27 explicit command migrations from the management node to one worker, with three repetitions for each combination of payload read size (0, 1, or 16 MiB) and minimum computation-loop duration (0, 0.25, or 2 seconds). An eight-second matrix-multiplication command occupies the source serial slot, and a second command waits for migration. Automatic migration is disabled in this measurement; the experiment driver invokes the reserve, claim, commit, and result-query operations explicitly.

The migrated shell command reads the prepared payload files and executes the elementwise workload with 262,144 elements and a 64-thread group. Payloads are written before timing begins.

Timing uses the controller's local elapsed-time clock. It starts immediately before lookup of the waiting command and ends when the driver observes the remote command's final result. It includes lookup, reservation, sandbox-process preparation, claim, commit, command execution, and result polling. Delivery to the source node, output retrieval, and the external numerical check follow this interval. Lookup and reservation together take a median of 0.818 seconds, with an observed range of 0.796--0.965 seconds across the 27 migrations. This stage includes sandbox-process preparation. Claim and commit take medians of 0.97 and 3.89 milliseconds, respectively.

Table~\ref{tab:overhead} reports the full timed migration path. With no additional payload, the medians are 1.86, 2.19, and 3.96 seconds for minimum loop durations of 0, 0.25, and 2 seconds. Reading 16 MiB raises them to 2.16, 2.51, and 4.20 seconds.

\input{artifacts/table_05/table.tex}

The approximately 1.86-second path at zero minimum loop duration makes migration costly for short commands. However, dynamic scheduling provides a net benefit in the concentrated two- and six-second batches, where idle remote capacity reduces source-node waiting.

\subsection{Static Agent Rollout Traces}
\label{sec:agent_rollout}
We collect sixteen mini-SWE-agent sessions through Harbor, initially assigning four independent sandbox instances to each of the four M4 nodes. The sessions use Step 5 Preview~\cite{step5preview} with mini-SWE-agent 2.4.6.
Each receives \code{metal\_nbody}: implementing a native Metal all-pairs gravitational simulation, an interactive renderer, a deterministic simulation CLI, and a benchmark entry point. The active execution budgets are 3,600 seconds per agent and 600 seconds per tool. Appendix~\ref{app:nbody_task} gives the complete task specification.
Static workspaces and \code{HOME} directories remain instance-local.
The task exercises source code generation, compilation, numerical testing, and repeated application runs.

Batch elapsed time runs from release of the sixteen sessions to the last sandbox stop; individual elapsed time runs from session start to sandbox stop.
Instrumentation distinguishes model requests, serial and parallel tools, tool waiting, final verification, and shutdown.
Model intervals include service waiting, reasoning, and generation.
Tool waiting spans command submission to its observed source start notification, including helper startup, communication, and queue delay; model-request time is separate.
Figure~\ref{fig:agent_timelines_combined} shows the measured static timelines alongside their dynamic-scheduling replay.

The measured batch lasts 67.6 minutes, with mean tool waiting of 21.0 minutes and mean individual elapsed time of 45.1 minutes.
These sixteen static trajectories contain 114 parallel and 1,165 serial tool commands.
These traces supply the fixed behavior for the scheduling replay below.

\paragraph{Sustained-throughput estimate.}
Subtracting each session's recorded tool waiting while retaining all other phase durations gives a mean elapsed time $T_0=24.12$ minutes.
The measured mean with four sessions per node is $T_4=45.12$ minutes.
If completed sessions are immediately replaced to maintain four active sessions per node, estimated per-node throughput increases from $1/T_0$ to $4/T_4$:
\[
S_{\mathrm{throughput}}=\frac{4/T_4}{1/T_0}
=\frac{4\times24.12}{45.12}\approx2.14.
\]
This compares one versus four sessions per node on the same hardware; the node count cancels.

\subsection{Trace-Driven Scheduling Replay}
\label{sec:replay}
Independent live runs can produce different model responses, command sequences, compilation failures, and execution durations. Comparing a live static run with a separate dynamic run would mix the scheduling effect with changes in agent behavior.
We therefore use paired trace-driven replay to compare both scheduling policies on the same measured trajectories.
The measured static instances use local workspaces. In the replay, recorded command paths are screened according to the migration-eligibility requirements in Section~\ref{sec:transfer}. With static scheduling, each replayed session finishes less than 1 ms from its measured finish time, and its total tool waiting differs by less than 8 ms. This validates the reconstruction of the recorded schedule before changing the policy.

Using migration delays measured in Section~\ref{sec:migration_overhead}, dynamic-scheduling replay reduces mean tool waiting from 21.0 to 17.1 minutes (18.8\%) and batch elapsed time from 67.6 to 52.5 minutes (22.3\%).

\input{artifacts/figure_09/combined.tex}
\input{artifacts/figure_08/figure.tex}

Figure~\ref{fig:agent_distribution} shows both the batch-level reduction and the distribution of individual elapsed times across the same sixteen trajectories.

The replay fixes model responses and their durations, command text and order, observed execution-slot occupancy, and local communication time. Each later submission waits for its preceding tool responses. The median delays measured in Section~\ref{sec:migration_overhead} are 0.818 seconds for lookup and reservation combined, 0.97 milliseconds for claim, and 3.89 milliseconds for commit; the median additional cleanup, communication, and result-polling delay is 0.935 seconds. The main replay uses these measured costs and a 0.015-second source-notification delay estimated as half the 0.030-second polling interval.

The main replay completes 276 command migrations.
Eight trajectories have less accumulated waiting and eight have more.
Mean waiting for the initially busiest owner, Mac mini \#A, falls from 32.7 to 15.8 minutes; \#B also improves, while \#C and \#D incur more waiting.
We vary migration overhead and scheduling parameters to assess the sensitivity of the replay results. Across the tested settings, the replay predicts reductions of 6.78--20.4\% in mean tool waiting and 15.6--27.8\% in batch elapsed time relative to measured static scheduling. Appendix~\ref{app:replay} details the 166 parameter settings and validation checks.

Appendix~\ref{app:agent_records} reports every paired trajectory and its elapsed time.

\section{Related Work}
\paragraph{Kernel evaluation and training.}
KernelBench couples correctness and performance evaluation with parallel generation and serialized GPU execution~\cite{kernelbench}. KernelGYM, developed for \mbox{Dr.\ Kernel}, supports distributed GPU evaluation, isolated subprocesses, and multi-turn reinforcement-learning data collection~\cite{drkernel}. Metal-Sci searches native Metal scientific kernels~\cite{metalsci}, while KernelArc coordinates optimization agents and shares validated results~\cite{kernelarc}. These systems schedule work at the granularity of candidate evaluations. \system{} schedules the individual tool commands of interactive agent sessions, which run in persistent native macOS instances with direct Apple GPU access.

\paragraph{Remote command execution.}
distcc distributes compilation by sending preprocessed sources and compiler arguments~\cite{distcc}. Bazel Remote Execution distributes build and test actions through a worker pool, transferring inputs and caching outputs in content-addressed storage~\cite{bazelremote}. These systems establish command-level remote execution. \system{} supports interactive tools in persistent native macOS instances, sharing workspace and \code{HOME} paths while preserving each agent's command-result interface.

\paragraph{Distributed scheduling.}
Sparrow delays task assignment until workers are ready to execute~\cite{sparrow}. Work stealing lets idle processors take ready computations from other processors~\cite{workstealing}. \system{} applies availability-based scheduling to waiting serial tool commands, using target reservations and source claims to coordinate migration while running processes remain local.

\paragraph{Agent interfaces and rollout environments.}
SWE-agent studies how command and observation interfaces shape automated software engineering~\cite{sweagent}. SkyRL-Agent uses asynchronous dispatch for multi-turn agent training and evaluation~\cite{skyrlagent}, and R2E-Gym provides executable software-engineering environments and hybrid verification~\cite{r2egym}. \system{} supplies native Apple Metal execution and command scheduling through Harbor, supporting Claude Code, Codex, and mini-SWE-agent in a common task and trajectory-collection workflow.

\paragraph{Accelerator sharing and execution isolation.}
Gandiva time-slices GPUs and migrates training jobs~\cite{gandiva}; Salus supports fast job switching and memory sharing~\cite{salus}. Firecracker isolates workloads with KVM-based microVMs~\cite{firecracker}, while gVisor implements a Linux application kernel to mediate system calls~\cite{gvisor}. \system{} uses Apple App Sandbox to restrict per-instance file access while retaining native macOS tools and Apple GPU access.

\section{Discussion and Limitations}

File-access restrictions apply to sandbox processes and their ordinary children; some application-launch paths remain outside this boundary. The system does not provide hostile multi-tenant containment or physical GPU exclusivity. Other applications and commands classified as parallel may still use the GPU.

Replay estimates scheduling effects while holding model behavior and phase durations fixed. It uses a source-notification delay estimated from the sandbox-process polling interval. It cannot predict changes in agent behavior caused by different tool feedback, or changes in kernel quality, training outcomes, or deployment reliability. Eight trajectories have less accumulated tool waiting and eight have more, even though mean tool waiting and batch elapsed time improve.

\section{Conclusion}
Metal Sandbox combines native Apple Metal execution, per-instance filesystem isolation, and command-level scheduling across interconnected Macs. Its Harbor integration supports Claude Code, Codex, and mini-SWE-agent through a shared task, verification, and trajectory-collection workflow. Four-node experiments demonstrate scaling under balanced submissions and throughput gains under concentrated submissions. Replaying fixed agent trajectories predicts reduced tool waiting and shorter batch completion under dynamic scheduling. Metal Sandbox provides execution infrastructure for training and evaluating models that generate Apple Metal kernels.

\renewcommand{\bibfont}{\raggedright}
\balance
\bibliographystyle{mlsys2025}
\bibliography{references}

\clearpage
\nobalance
\appendix
\section{Static Rollout and Replay Records}
\label{app:agent_records}
The measured static batch and its dynamic scheduling replay retain the same sixteen trajectories.
Indices in Table~\ref{tab:agent_individual} identify paired trajectories with fixed model responses and command behavior.
Mac mini labels \#A--\#D identify the initial owner throughout both schedules.
Figure~\ref{fig:agent_distribution} compares batch elapsed time and individual duration distributions.

\input{artifacts/table_07/table.tex}

\section{Scheduling Replay Details}
\label{app:replay}
The replay retains all sixteen measured trajectories, including five tools without observed start notifications. No new model responses or candidate executions are produced. All identifiers and owner-node labels match Appendix~\ref{app:agent_records}.

\paragraph{Trace coverage and timing.}
Inputs contain 1,275 complete model phases, 1,279 tool calls (1,165 serial and 114 parallel), and 144 setup or verification commands. All 1,274 tools with execution-start records have matching command-argument hashes. Responses, command order, observed execution-slot occupancy, and local communication durations remain fixed. Each subsequent model request, tool call, verification step, or shutdown is delayed until its preceding responses are available. Tool waiting ends at the source start notification, which may arrive after remote execution begins.

\paragraph{Scheduling and portability.}
Each node has one serial execution slot and a first-in, first-out (FIFO) queue. Every second, the controller checks source nodes in node-ID order. A target node must have a free serial slot and no local waiting serial commands; selection uses current instance count, then node ID. Waiting commands are selected by command-ID order, without inspecting future durations. Reservation occupies the target slot; a source-local start before claim cancels migration. Commands already running are not migrated; agent sessions, parallel tools, setup, and verification remain on their owner nodes.

The replay filters the recorded command text for absolute paths to owner-local files, leaving 1,131 serial tools eligible for migration. Environment-variable references to node-private dependencies require portable environments for these predictions.

\paragraph{Migration costs.}
The 27 measured migrations provide median combined lookup and reservation, claim, and commit delays of 0.818, 0.00097, and 0.00389 seconds. A further 0.935 seconds is the median difference between remote-result observation time and candidate-process execution time; it includes cleanup, communication, and result polling. Source notification adds an estimated 0.015 seconds, half the configured 0.030-second sandbox-process polling interval. The main replay uses overhead scale 1, controller start offset 0, and command-ID ordering seed 0.

\paragraph{Validation and sensitivity.}
With static scheduling, each replayed session finishes less than 1 ms from its measured finish time, and its total tool waiting differs by less than 8 ms. Validation also checks phase dependencies and durations, keeps parallel commands local, and enforces migration eligibility, one serial command per node, target reservations, and selection based only on current node status. The sensitivity study combines four migration-cost scales \{0.5,1,2,4\}, four controller start offsets \{0,0.25,0.5,0.75\} seconds, and ten command-ID ordering seeds 0--9, giving $4\times4\times10=160$ configurations. Six additional configurations vary only source-notification delay \{0,0.03,0.1,0.5\} seconds or controller-check interval \{2,5\} seconds from the main settings. All 166 pass the scheduling checks. Mean tool waiting spans 16.7--19.6 minutes; batch elapsed time spans 48.8--57.1 minutes. All configurations and trajectories are retained in the artifact.
\onecolumn
\section{Default Parallel Allowlist}
\label{app:allowlist}
\input{artifacts/command_allowlist/appendix.tex}

\section{Task Specification: \textnormal{\code{metal\_nbody}}}
\label{app:nbody_task}
\input{artifacts/task_nbody/appendix.tex}

\end{document}

%% file: artifacts/figure_01/figure.tex
\begin{figure}[t]
\centering
\includegraphics[width=\columnwidth]{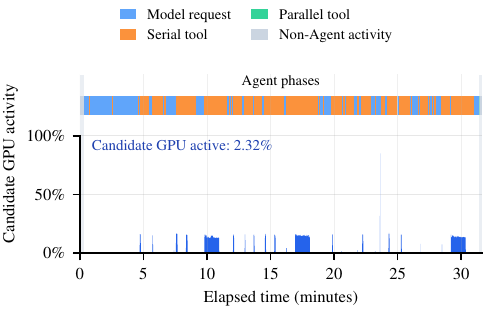}
\caption{One \code{metal\_nbody} agent session without competing sessions. GPU activity totals 44.0 seconds (2.32\%) over 31.6 minutes. Agent phases and GPU activity share the session timeline. The GPU curve is the active fraction of each one-second window; its area gives the total active time. Gray denotes activity outside the recorded model and tool phases.}
\label{fig:single_agent_motivation}
\end{figure}

%% file: artifacts/table_01/table.tex
\begin{table}[t]
\centering\small
\renewcommand{\arraystretch}{1.12}
\setlength{\tabcolsep}{0pt}
\caption{Execution environments for Metal agents. $\checkmark$: supported; $\times$: unsupported; $\triangle$: partial. Docker denotes standard Linux containers; macOS VMs expose virtual Metal graphics.}
\label{tab:execution_choices}
\vskip 0.15in
\begin{tabular}{@{}>{\raggedright\arraybackslash}m{0.40\columnwidth}*{4}{>{\centering\arraybackslash}m{0.15\columnwidth}}@{}}
\toprule
Capability & Docker & \shortstack{macOS\\VM} & \shortstack{bare\\macOS} & \shortstack{\textbf{Metal}\\\textbf{Sandbox}} \\
\midrule
macOS programs & $\times$ & $\checkmark$ & $\checkmark$ & $\pmb{\checkmark}$ \\
Host OS access & $\times$ & $\times$ & $\checkmark$ & $\pmb{\checkmark}$ \\
Filesystem isolation & $\checkmark$ & $\checkmark$ & $\times$ & $\pmb{\checkmark}$ \\
Native Apple GPU access & $\times$ & $\triangle$ & $\checkmark$ & $\pmb{\checkmark}$ \\
GPU timing coordination & $\times$ & $\times$ & $\times$ & $\pmb{\checkmark}$ \\
Cross-Mac scheduling & $\times$ & $\times$ & $\times$ & $\pmb{\checkmark}$ \\
\bottomrule
\end{tabular}
\vskip -0.1in
\end{table}

%% file: artifacts/native_command_example/example.tex
\noindent\fbox{%
\begin{minipage}{\dimexpr\columnwidth-2\fboxsep-2\fboxrule\relax}
\raggedright\small
{\footnotesize\ttfamily
\$ mkdir \textasciitilde/demo\par
\textcolor{red}{[ALLOWED; exit 0]}\par\medskip
\$ mkdir /Users/sandbox1/demo\par
\textcolor{red}{[DENIED; exit 1]}\par
mkdir: /Users/sandbox1/demo:\par
\hspace*{1em}Operation not permitted\par
}
\end{minipage}%
}\par\smallskip

%% file: artifacts/figure_02/figure.tex
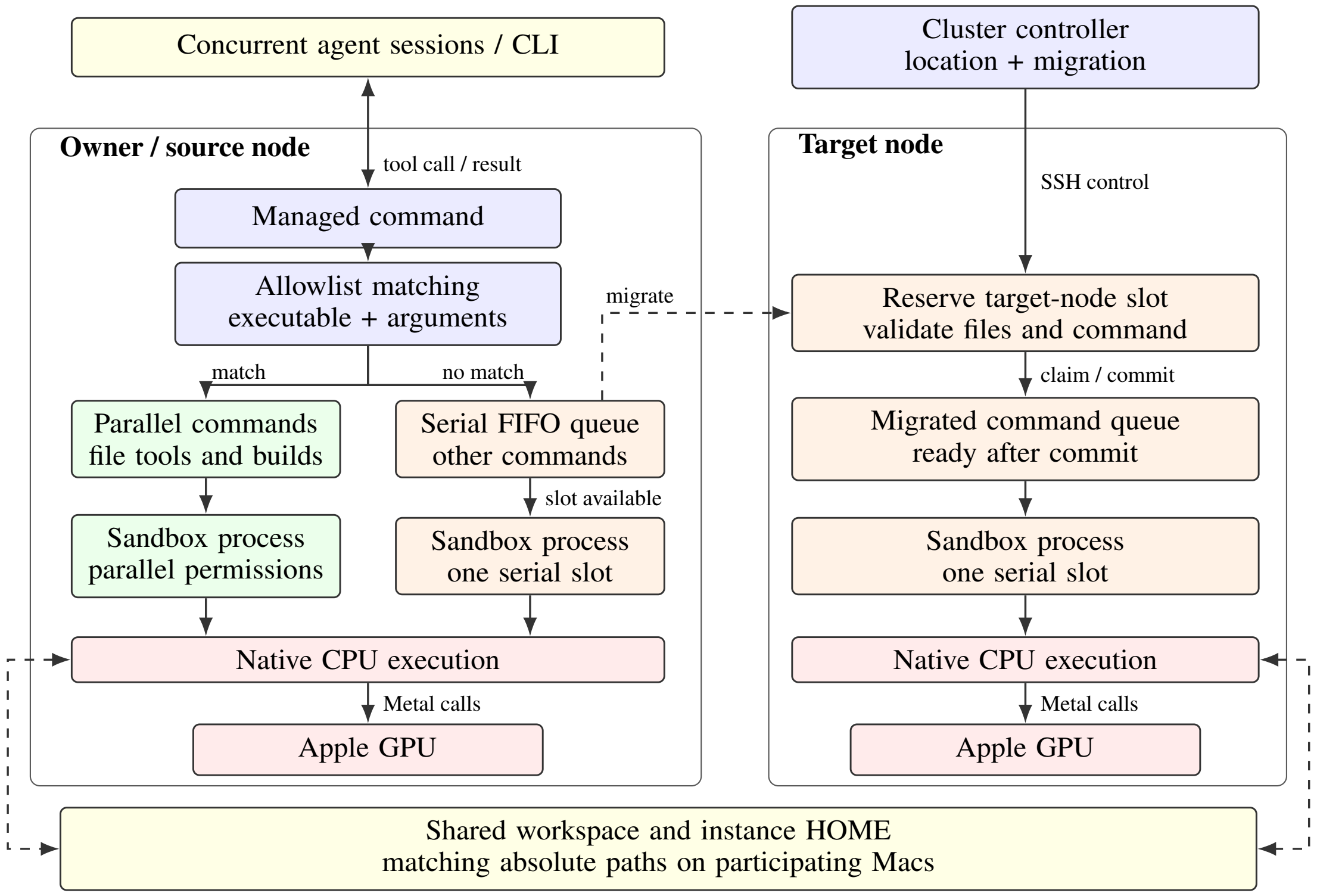
\begin{figure*}[t]
\centering
\input{figures/figure_02_architecture.tex}
\caption{Command scheduling and execution. Allowlist matches receive parallel execution permissions; other commands enter the node's serial FIFO queue. Both paths execute native commands in an App Sandbox process, and Metal calls submit work to the Apple GPU. GPU timing commands use the serial path under the configured rules. Eligible waiting commands can migrate through target-node reservation, claim, and commit, using shared instance files. Outputs return to the original tool call.}
\label{fig:architecture}
\end{figure*}

%% file: figures/figure_02_architecture.tex
\begingroup
\definecolor{archblue}{HTML}{EBEBFF}
\definecolor{archgreen}{HTML}{EBFFEB}
\definecolor{archyellow}{HTML}{FFFFE6}
\definecolor{archorange}{HTML}{FFF2E6}
\definecolor{archred}{HTML}{FFEBEB}
\begin{tikzpicture}[font=\footnotesize,>=Latex,
  box/.style={draw=black!80,line width=0.55pt,rounded corners=2pt,align=center,minimum height=0.65cm,inner sep=4pt},
  flow/.style={->,draw=black!80,line width=0.65pt},
  note/.style={font=\scriptsize,inner sep=1.5pt}]
\draw[draw=black!65,rounded corners=3pt] (-0.05,-0.55) rectangle (7.4,-7.85);
\draw[draw=black!65,rounded corners=3pt] (8.15,-0.55) rectangle (13.9,-7.85);
\node[anchor=west,font=\footnotesize\bfseries,inner sep=2pt] at (0.2,-0.75) {Owner / source node};
\node[anchor=west,font=\footnotesize\bfseries,inner sep=2pt] at (8.4,-0.75) {Target node};
\node[box,fill=archyellow,text width=6.3cm] (agents) at (3.7,0.35) {Concurrent agent sessions / CLI};
\node[box,fill=archblue,text width=4.9cm] (controller) at (11,0.35) {Cluster controller\\location + migration};
\node[box,fill=archblue,text width=4.0cm] (helper) at (3.7,-1.55) {Managed command};
\node[box,fill=archblue,text width=4.0cm] (filter) at (3.7,-2.5) {Allowlist matching\\executable + arguments};
\node[box,fill=archgreen,text width=2.7cm] (parallel) at (1.9,-4.0) {Parallel commands\\file tools and builds};
\node[box,fill=archorange,text width=2.7cm] (queue) at (5.5,-4.0) {Serial FIFO queue\\other commands};
\node[box,fill=archgreen,text width=2.7cm] (parallelrun) at (1.9,-5.3) {Sandbox process\\parallel permissions};
\node[box,fill=archorange,text width=2.7cm] (serialrun) at (5.5,-5.3) {Sandbox process\\one serial slot};
\node[box,fill=archred,text width=6.3cm,minimum height=0.5cm] (cpu) at (3.7,-6.45) {Native CPU execution};
\node[box,fill=archred,text width=3.6cm,minimum height=0.5cm] (gpu) at (3.7,-7.45) {Apple GPU};
\node[box,fill=archorange,text width=4.9cm] (reservation) at (11,-2.6) {Reserve target-node slot\\validate files and command};
\node[box,fill=archorange,text width=4.9cm] (targetqueue) at (11,-4.0) {Migrated command queue\\ready after commit};
\node[box,fill=archorange,text width=4.9cm] (targetrun) at (11,-5.3) {Sandbox process\\one serial slot};
\node[box,fill=archred,text width=4.9cm,minimum height=0.5cm] (targetcpu) at (11,-6.45) {Native CPU execution};
\node[box,fill=archred,text width=3.6cm,minimum height=0.5cm] (targetgpu) at (11,-7.45) {Apple GPU};
\node[box,fill=archyellow,text width=13.0cm] (storage) at (6.925,-8.55) {Shared workspace and instance HOME\\matching absolute paths on participating Macs};
\draw[flow,<->] (agents)--node[note,right=3pt,pos=0.78]{tool call / result}(helper);
\draw[flow] (helper)--(filter);
\draw[flow] (filter.south)--(3.7,-3.4)--node[note,above right,pos=1]{match}(1.9,-3.4)--(parallel.north);
\draw[flow] (filter.south)--(3.7,-3.4)--node[note,above left,pos=1]{no match}(5.5,-3.4)--(queue.north);
\draw[flow] (parallel)--(parallelrun);
\draw[flow] (queue)--node[note,right=3pt]{slot available}(serialrun);
\draw[flow] (parallelrun.south)--(parallelrun.south |- cpu.north);
\draw[flow] (serialrun.south)--(serialrun.south |- cpu.north);
\draw[flow] (cpu)--node[note,right=3pt]{Metal calls}(gpu);
\draw[flow] (controller)--node[note,right=3pt]{SSH control}(reservation);
\draw[flow,dashed] ([xshift=0.8cm]queue.north)--(6.3,-2.6)--node[note,above,pos=0.2]{migrate}(reservation.west);
\draw[flow] (reservation)--node[note,right=3pt]{claim / commit}(targetqueue);
\draw[flow] (targetqueue)--(targetrun);
\draw[flow] (targetrun)--(targetcpu);
\draw[flow] (targetcpu)--node[note,right=3pt]{Metal calls}(targetgpu);
\draw[flow,<->,dashed] (cpu.west)--(-0.3,-6.45)--(-0.3,-8.55)--(storage.west);
\draw[flow,<->,dashed] (targetcpu.east)--(14.15,-6.45)--(14.15,-8.55)--(storage.east);
\end{tikzpicture}
\endgroup

%% file: artifacts/figure_03/figure.tex
\begin{figure}[t]
\centering
\includegraphics[width=\columnwidth]{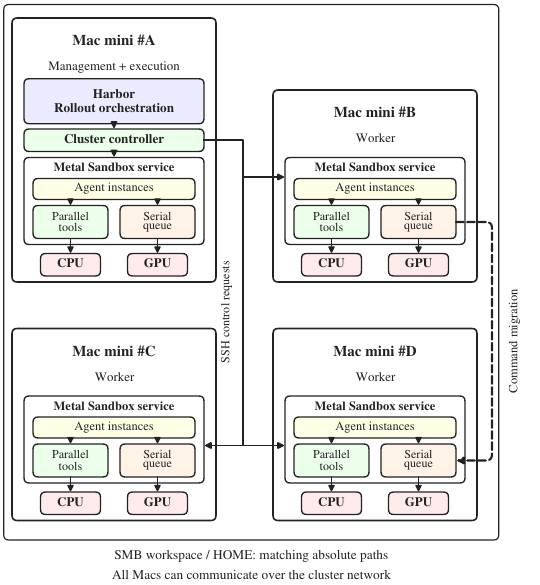}
\caption{Four-Mac deployment. Harbor and the cluster controller run on Mac mini \#A, which also executes work. Solid arrows show SSH control requests; the dashed arrow shows command migration from \#B's waiting queue to \#D. CPU/GPU arrows show parallel preparation and serial Metal execution. The agent stays on its owner Mac. All nodes have network access and prepared SMB workspace/\code{HOME} paths.}
\label{fig:cluster_deployment}
\end{figure}

%% file: artifacts/figure_04/figure.tex
\begin{figure}[t]
\centering
\input{figures/figure_04_transfer_protocol.tex}
\caption{Command migration through reservation, claim, and commit. The source-node claim prevents local execution before the target node starts the command.}
\label{fig:protocol}
\end{figure}
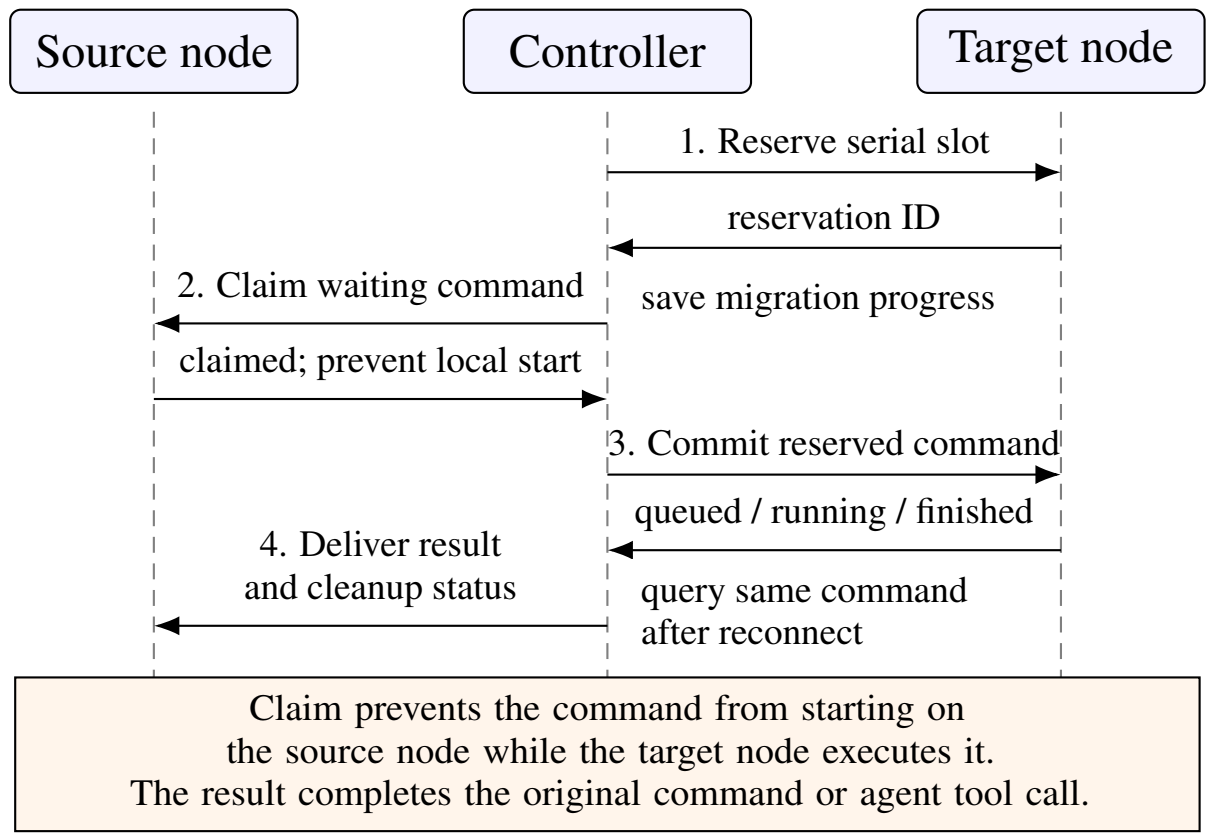

%% file: figures/figure_04_transfer_protocol.tex
\begin{tikzpicture}[x=1cm,y=1cm,font=\small,>=Latex]
\foreach \xx/\name in {0/Source node,3/Controller,6/Target node} {
  \node[draw,rounded corners=2pt,fill=blue!5,minimum width=1.9cm,minimum height=0.55cm] at (\xx,0) {\name};
  \draw[densely dashed,gray] (\xx,-0.4)--(\xx,-4.15);
}
\draw[->] (3,-0.8)--node[above,font=\scriptsize]{1. Reserve serial slot}(6,-0.8);
\draw[->] (6,-1.3)--node[above,font=\scriptsize]{reservation ID}(3,-1.3);
\node[anchor=west,font=\scriptsize,fill=white] at (3.10,-1.65) {save migration progress};
\draw[->] (3,-1.8)--node[above,font=\scriptsize]{2. Claim waiting command}(0,-1.8);
\draw[->] (0,-2.3)--node[above,font=\scriptsize]{claimed; prevent local start}(3,-2.3);
\draw[->] (3,-2.8)--node[above,font=\scriptsize]{3. Commit reserved command}(6,-2.8);
\draw[->] (6,-3.3)--node[above,font=\scriptsize]{queued / running / finished}(3,-3.3);
\node[anchor=west,align=left,font=\scriptsize,fill=white] at (3.10,-3.7) {query same command\\after reconnect};
\draw[->] (3,-3.8)--node[above,align=center,font=\scriptsize]{4. Deliver result\\and cleanup status}(0,-3.8);
\node[draw,align=center,font=\scriptsize,text width=7.6cm,fill=orange!8] at (3,-4.65) {Claim prevents the command from starting on the source node while the target node executes it.\\The result completes the original command or agent tool call.};
\end{tikzpicture}

%% file: artifacts/table_03/table.tex
\begin{table}[t]
\centering\small
\renewcommand{\arraystretch}{1.15}
\caption{Fixed scheduling workloads. TG denotes the requested threads per threadgroup, bounded by the pipeline limit.}
\label{tab:workloads}
\vskip 0.15in
\begin{tabular}{@{}>{\raggedright\arraybackslash}p{0.26\columnwidth}@{\hspace{6pt}}>{\raggedright\arraybackslash}p{0.58\columnwidth}@{\hspace{6pt}}r@{}}
\toprule
Kernel & Input and computation & TG \\
\midrule
Matmul & Two $128\times128$ matrices; one output element/thread & 64 \\
Convolution & $256\times256$, single channel; $3\times3$ stencil, zero boundary & 256 \\
Elementwise & 262,144 elements; $a+b$ & 64 \\
Reduction & 1,024 groups of 256; sum per group & 256 \\
Normalization & 1,024 groups of 256; mean/variance, $\epsilon=10^{-5}$ & 64 \\
Fused & 262,144 elements; $\max(0,ab+0.125)$ & 256 \\
\bottomrule
\end{tabular}
\vskip -0.1in
\end{table}

%% file: artifacts/table_04/table.tex
\begin{table*}[t]
\centering\small
\caption{Twelve commands per batch with two-second minimum computation-loop durations. Times are medians [minimum, maximum] over three independent batches. Throughput is the median number of commands per hour calculated from each batch. All 288 reported executions pass the numerical check.}
\label{tab:throughput}
\vskip 0.15in
\begin{tabular}{@{}rlrrrr@{}}
\toprule
Nodes & Submission & Static time (s) & Dynamic time (s) & Static/hour & Dynamic/hour \\
\midrule
1 & Balanced & 25.2 [25.0, 25.2] & 25.1 [25.1, 25.2] & 1,710 & 1,720 \\
2 & Balanced & 12.9 [12.8, 13.0] & 12.8 [12.8, 12.9] & 3,350 & 3,380 \\
4 & Balanced & 6.62 [6.62, 6.69] & 6.61 [6.60, 6.65] & 6,520 & 6,540 \\
4 & Concentrated & 25.3 [25.2, 25.3] & 15.1 [14.0, 16.9] & 1,710 & 2,870 \\
\bottomrule
\end{tabular}
\vskip -0.1in
\end{table*}

%% file: artifacts/figure_05/figure.tex
\begin{figure}[t]
\centering
\includegraphics[width=\columnwidth]{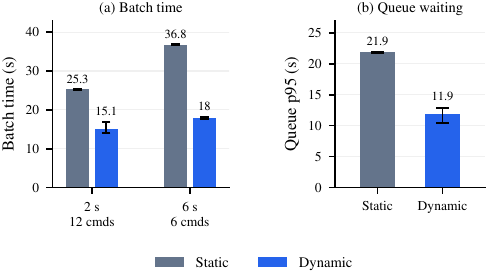}
\caption{Concentrated submission to one node in a four-Mac cluster. Panel (a) shows batch time for two-second loops with twelve commands and six-second loops with six commands. Panel (b) shows within-batch queue p95 for the two-second workload. Bars show medians and whiskers show the minimum--maximum range of three independent batches.}
\label{fig:throughput}
\end{figure}

%% file: artifacts/table_05/table.tex
\begin{table}[t]
\centering\small
\caption{Median explicit command migration time (seconds) from lookup of the waiting command to its remote result. Columns give minimum computation-loop durations. Each cell summarizes three repetitions. A zero duration retains warmup and GPU sampling.}
\label{tab:overhead}
\vskip 0.15in
\begin{tabular}{@{}rrrr@{}}
\toprule
& \multicolumn{3}{c}{Minimum loop duration (s)} \\
Read (MiB) & 0 & 0.25 & 2 \\
\midrule
0 & 1.86 & 2.19 & 3.96 \\
1 & 1.92 & 2.13 & 3.95 \\
16 & 2.16 & 2.51 & 4.20 \\
\bottomrule
\end{tabular}
\vskip -0.1in
\end{table}

%% file: artifacts/figure_09/combined.tex
\begin{figure*}[!t]
\centering
\begin{minipage}{\textwidth}
\centering
{\small (a) Measured static scheduling\par}
\vspace{2pt}
\includegraphics[width=\textwidth]{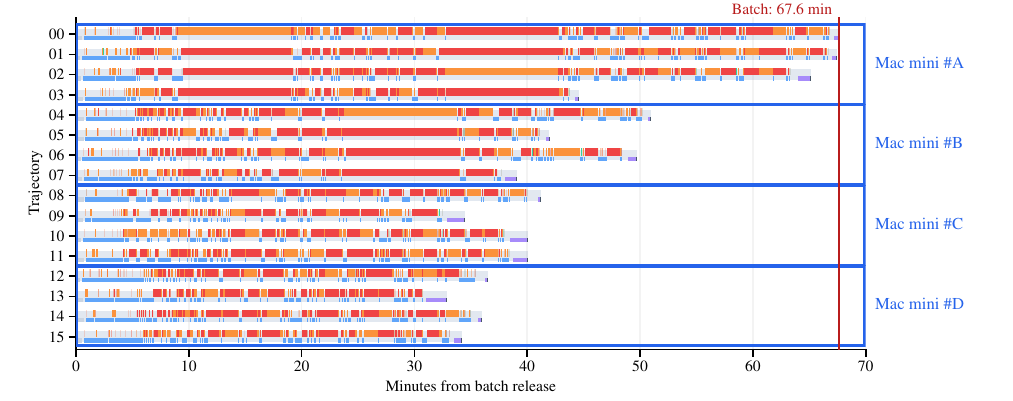}
\end{minipage}
\par\vspace{5pt}
\begin{minipage}{\textwidth}
\centering
{\small (b) Dynamic-scheduling replay\par}
\vspace{2pt}
\includegraphics[width=\textwidth]{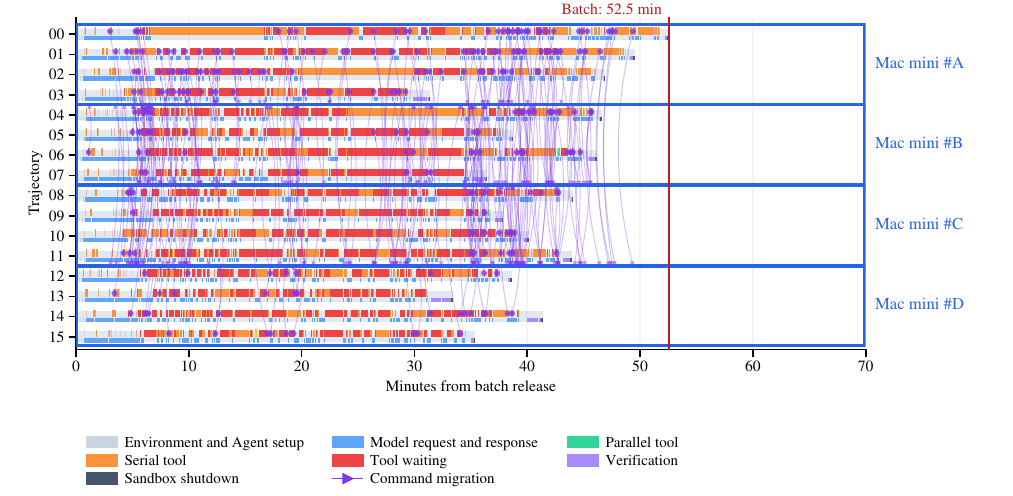}
\end{minipage}
\caption{Sixteen agent timelines under (a) measured static scheduling and (b) dynamic-scheduling replay. Red lines mark batch elapsed time, when the last sandbox stops. Blue outlines group four sessions per owner Mac mini, \#A--\#D.}
\label{fig:agent_timelines_combined}
\end{figure*}

%% file: artifacts/figure_08/figure.tex
\begin{figure}[htbp]
\centering
\includegraphics[width=\columnwidth]{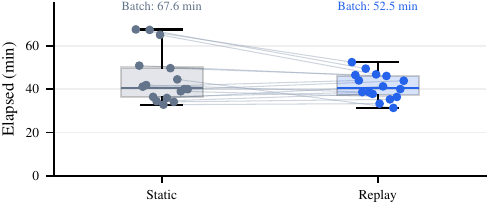}
\caption{Batch elapsed time and individual duration distributions for the measured static batch and its dynamic scheduling replay. Labels give batch elapsed time; circles and connecting lines retain all sixteen paired trajectories. Replay holds model behavior and phase durations fixed.}
\label{fig:agent_distribution}
\end{figure}

%% file: artifacts/table_07/table.tex
\begin{table}[htbp]
\centering\small
\setlength{\tabcolsep}{4pt}
\renewcommand{\arraystretch}{1.05}
\caption{Individual elapsed time and accumulated tool waiting in minutes for every paired trajectory. Elapsed time runs from session start to sandbox stop. Replay fixes the measured model responses and phase durations under the assumptions in Appendix~\ref{app:replay}.}
\label{tab:agent_individual}
\vskip 0.15in
\begin{tabular}{crrrr}
\toprule
& \multicolumn{2}{c}{Measured static} & \multicolumn{2}{c}{Dynamic replay} \\
Trace / Mac & Elapsed & Waiting & Elapsed & Waiting \\
\midrule
00 / \#A & 67.6 & 29.0 & 52.5 & 13.1 \\
01 / \#A & 67.4 & 40.5 & 49.6 & 21.6 \\
02 / \#A & 65.1 & 32.6 & 46.9 & 14.7 \\
03 / \#A & 44.6 & 28.7 & 31.4 & 13.9 \\
04 / \#B & 50.9 & 18.2 & 46.6 & 13.5 \\
05 / \#B & 42.0 & 22.4 & 38.7 & 18.9 \\
06 / \#B & 49.8 & 25.5 & 46.1 & 21.6 \\
07 / \#B & 39.1 & 20.9 & 36.5 & 17.9 \\
08 / \#C & 41.2 & 18.5 & 44.1 & 21.2 \\
09 / \#C & 34.5 & 15.0 & 37.9 & 19.2 \\
10 / \#C & 40.1 & 15.3 & 40.1 & 16.3 \\
11 / \#C & 40.0 & 16.2 & 43.9 & 20.4 \\
12 / \#D & 36.5 & 12.8 & 38.6 & 14.8 \\
13 / \#D & 32.9 & 13.2 & 33.4 & 13.4 \\
14 / \#D & 36.0 & 15.0 & 41.3 & 19.0 \\
15 / \#D & 34.2 & 12.0 & 35.3 & 13.3 \\
\midrule
Mean & 45.1 & 21.0 & 41.4 & 17.1 \\
\bottomrule
\end{tabular}
\vskip -0.1in
\end{table}

%% file: artifacts/command_allowlist/appendix.tex
The following is the complete shipped parallel allowlist template. Each name in the first two groups denotes the executable at the stated directory. The final group denotes paths resolved from the installed development tools when an instance is created. All entries accept any argument prefix except CMake, which requires \code{-{}-build}. The administrator can change this scheduling policy; the list classifies commands and does not detect GPU use.

\begin{table}[h]
\centering\small
\renewcommand{\arraystretch}{1.2}
\begin{tabular}{@{}>{\raggedright\arraybackslash}p{0.17\columnwidth}>{\raggedright\arraybackslash}p{0.79\columnwidth}@{}}
\toprule
Executable directory & Complete entries \\
\midrule
\code{/bin/} & \code{ls}, \code{cat}, \code{pwd}, \code{sleep}, \code{echo}, \code{date}, \code{test}, \code{mkdir}, \code{rmdir}, \code{cp}, \code{mv}, \code{rm}, \code{ln}, \code{chmod}, \code{df}, \code{realpath} \\
\code{/usr/bin/} & \code{touch}, \code{head}, \code{tail}, \code{grep}, \code{egrep}, \code{fgrep}, \code{sed}, \code{awk}, \code{sort}, \code{uniq}, \code{wc}, \code{cut}, \code{tr}, \code{tee}, \code{diff}, \code{cmp}, \code{comm}, \code{paste}, \code{join}, \code{basename}, \code{dirname}, \code{readlink}, \code{stat}, \code{file}, \code{du}, \code{find}, \code{jq}, \code{git}, \code{curl}, \code{tar}, \code{zip}, \code{unzip}, \code{gzip}, \code{gunzip}, \code{shasum}, \code{printf}, \code{true}, \code{false}, \code{make} \\
Installed tools & C compiler, C++ compiler, Xcode Git, Xcode Make, Ninja, Metal compiler, \code{metallib}, and CMake with argument prefix \code{-{}-build} \\
\bottomrule
\end{tabular}
\end{table}

%% file: artifacts/task_nbody/appendix.tex
\begingroup
\setlength{\FrameRule}{0.4pt}
\setlength{\FrameSep}{8pt}
\renewcommand{\paragraph}[1]{\par\medskip\noindent#1\par\nobreak\smallskip}
\begin{framed}
\small\normalfont
\setlength{\parindent}{0pt}
\paragraph{Execution and submission.}
This is a complete project task to be completed autonomously. You receive the full specification and public attachments at the start. Implement, build, test, and fix the project within the same session without waiting for clarification, approval, or further evaluation feedback. Decide unspecified details yourself and document them in the README. Tool calls are unlimited. Submit complete source code, dependency configuration, a launch entry point, and usage instructions; do not submit plans, placeholder modules, or unfinished implementations. The project is frozen after final submission or the time limit and evaluated centrally, without returning errors for further repair. Interactions described below are features of the finished product and do not require user participation during development.

\paragraph{L: Star Cluster Workbench---Metal particle simulation and graphics optimization.}
A science museum wants to display dynamic structures formed by particles under mutual gravitational attraction. The exhibit must run continuously without frequent stalls when visitors move the camera, pause, or regenerate the cluster. Build a native interactive macOS application connecting computational correctness, GPU scheduling, and visual presentation. Research-grade astrophysical accuracy is unnecessary, but prerecorded video, regular rotation, or a small particle set cannot substitute for the specified interactions. The goal is a graphics implementation that can be repeatedly checked using common inputs, numerical comparisons, and device measurements.

The display may begin with a simple small-particle state and then switch to a dense cluster. Both scales use the same computational definition; do not use a separate approximate simulation for the display. This sequence is a usage suggestion, not a requirement for a preset-scene library. Graphics optimization may change data organization and execution, but must preserve the physical meaning of exported results. Numerical checks compare short deterministic runs; long-term visual trajectory differences are not error thresholds. Choose the camera, particle colors, and background to make spatial structure clear.

Deliver an application that performs three-dimensional all-pairs gravitational computation and rendering with Metal, plus a terminal benchmark entry point. The program must load external state, advance with a fixed time step, export results, and measure performance. Establish small-scale correctness before optimizing to sixty frames per second on the specified device and particle count. Choose resource organization and scheduling yourself; evaluation uses correct output and measured frame times under the same workload.

\paragraph{Working environment and self-testing.}
The current directory is the project workspace prepared by Harbor inside Metal Sandbox. Keep source code, builds, and outputs in this writable directory, without relying on the host task-package path. Keep every candidate build and run inside the management service's App Sandbox. Do not bypass restrictions through execution outside the sandbox, system services, elevated privileges, or host configuration changes.

The target is macOS on a physical Apple Silicon GPU. Use the actual clang, clang++, swift/swiftc, metal, and metallib resolved by the runtime PATH, together with the supplied CC, CXX, and SDKROOT. Check available tools and preserve the existing environment. System xcrun cannot run here; do not use it to locate tools or SDKs or assume an xcodebuild project is available. You may compile and link native frameworks directly, compile separate shaders with metal/metallib, or compile Metal source at runtime. Record missing tools honestly and do not install them on the host.

The terminal is not a PTY, so commands must not require interactive confirmation. If a bench-server is needed, communicate with it through child-process pipes from a runner in the same managed execution, without relying on a console to keep stdin open. GUI operation, visible windows, screenshots, and presentation timing remain to be verified in this environment; retain the application and interactions. If restricted, report actual errors and unverified items rather than claiming an offscreen substitute is complete. Numerical commands must be independent of windows. The performance target is a Mac mini M4 and 32 GiB memory. Record the actual machine, system, tools, and available measurement capabilities. On unmatched hardware or without exclusive access, report exploratory results without claiming the target is met. Multiple-instance scheduling does not guarantee GPU exclusivity. Evaluators calibrate formal measurements under suitable conditions; application-reported metrics are not the sole evidence.

\paragraph{M1: Project and entry points.}
Deliver a native macOS project, README, and executable \code{run.sh}. Support \code{./run.sh app}, \code{./run.sh simulate -{}-input FILE -{}-steps N -{}-output FILE}, and \code{./run.sh benchmark -{}-input FILE -{}-output DIR}. The simulation entry point must operate without mouse interaction, read external files, and write parseable JSON. Report actual build logs on failure.

\paragraph{M2: Inputs and scale.}
The input object contains \code{dt}, \code{epsilon}, and \code{particles}. Each particle is \code{\{id, mass, position:[x,y,z], velocity:[vx,vy,vz]\}}. IDs are unique signed 32-bit integers; mass lies in $[0.001,10]$, absolute position components do not exceed 100, and absolute velocity components do not exceed 10. The ranges are $dt\in[0.0001,0.02]$ and $epsilon\in[0.05,1]$; every number is finite. Support 2--8,192 particles. Invalid inputs or nonfinite computed values must produce a nonzero exit and an explanation.

\paragraph{M3: Acceleration definition.}
Use $G=1$. For every particle $i$,
\[
 a_i=\sum_{j\ne i}\frac{m_j(p_j-p_i)}{(\|p_j-p_i\|^2+epsilon^2)^{3/2}}.
\]
All accelerations in a step read positions from the start of that step; do not mix sequentially updated positions into the same step. Always include softening. Coincident particles must not produce NaN.

\paragraph{M4: Integration and export.}
Use semi-implicit Euler: $v_{new}=v_{old}+a\,dt$ and $p_{new}=p_{old}+v_{new}\,dt$. $N$ is an integer from 0 to 1,000. The simulate command advances exactly $N$ steps and outputs \code{\{dt,epsilon,steps:N,particles\}}, retaining each particle's \code{id}, \code{mass}, \code{position}, and \code{velocity} in input order. For small-scale tests after one and ten steps, each component's error must be at most $10^{-4}+10^{-3}|\text{reference}|$. Do not judge correctness using chaotic divergence of long-term trajectories.

\paragraph{M5: GPU implementation.}
All-pairs interaction and integration in every step must execute in Metal compute shaders; rendering must also use Metal. Do not substitute a CPU main loop, reduce the particles participating in interactions, cache fixed-input trajectories, or replace the formula with approximate neighborhood truncation. The CPU may handle input, UI, and resource management.

\paragraph{M6: Display and camera.}
Display every particle by default. Support mouse rotation, scroll-wheel zoom, and restoration of the default camera. Actual position updates drive the display. Particles may be points or small quads; complex materials are unnecessary. Camera changes must not change physical state, and window resizing must not stretch the coordinate space.

\paragraph{M7: Pause and reset.}
Provide play/pause, single-step, reset, load, and current-state export controls, with the same export format as simulate. Show particle count and cumulative step count. Paused state remains unchanged; single-step advances exactly once; reset restores the most recently loaded initial state and zero steps. Interactive mode advances one fixed-$dt$ step per rendered frame, without adapting the step size to wall time.

\paragraph{M8: Benchmark workload.}
The benchmark uses an evaluator-supplied state of 8,192 particles, $dt=1/120$, and $epsilon=0.05$. Render at $1920\times1080$ physical pixels. Each frame advances exactly one step and draws all particles. Dynamic resolution, skipped simulation steps, and reduced particle count are prohibited. Keep the camera fixed and the window visible during benchmarking.

\paragraph{M9: Performance target.}
After a ten-second warmup, measure continuously for sixty seconds, with one task exclusively using the device and a 60 Hz display configuration. Average presented frame rate must be at least 59, presentation-interval p95 must not exceed 20 ms, and no stall may exceed 200 ms. This tolerance expresses a stable sixty-frame-per-second target; it does not require every frame to be strictly below 16.67 ms. Numerical correctness and workload checks must pass before performance is assessed.

\paragraph{M10: Measurement artifacts.}
The benchmark exports per-frame submission, GPU completion, and actual presentation timestamps, step number, particle count, drawable size, and environment information. Explicitly mark unsupported timing fields as null and explain them. Do not pass off CPU submission time as GPU execution time. Trusted evaluation separately collects presentation evidence; application reports are not the only basis for assessment.

\paragraph{M11: Errors and resources.}
Load states with 2, 17, and 8,192 particles in sequence and switch between paused and running states ten times. There must be no out-of-bounds accesses, crashes, or stale-buffer artifacts. Preserve a usable interface and show a message on invalid input. Document self-test commands, known limitations, and one actual before/after optimization measurement in the README. Do not invent unexecuted results.

\paragraph{Technical scope.}
Use Swift, Objective-C, or C++ for host code and Metal Shading Language for GPU algorithms. Choose a native window framework. Do not substitute browser rendering, CUDA, external solver services, or video playback. Floating-point precision is your choice, but the specified error bounds must hold. Do not emit preset answers based on evaluation filenames.

Collision merging, gravitational tree approximations, multiple GPUs, physical cosmological units, long-term orbital stability, model imports, and complex postprocessing are out of scope. A realistic-looking nebula does not replace formula verification. The same frame rate is not required on Linux or different Mac models.

\paragraph{Suggested but optional.}
The following are optional and do not affect the mandatory requirements:
\begin{enumerate}
\item[B1:] Improve particle depth and background composition without reducing work.
\item[B2:] Provide an analysis panel explaining numerical errors and measurements.
\item[B3:] Improve camera reset, input feedback, and keyboard operation.
\item[B4:] Explain a real bottleneck and evidence for its resolution.
\end{enumerate}

\paragraph{Your decisions.}
Choose thread organization, memory layout, buffer reuse, compute/render synchronization, particle shape, and colors. No particular technique is required, but explain what the exported timing metrics mean. If time is limited, prioritize small-scale numerical correctness and reproducible entry points. Extra visual effects must not reduce benchmark work.

\paragraph{Delivery instructions.}
Document the shortest steps from a clean directory to the main interface, including required configuration, default demonstration data, and reset behavior. Include source code, dependency versions, and required local resources; exclude credentials and personal information. Invalid input must produce understandable feedback rather than silently changing a previously correct state. Record only tests actually performed, explicitly identify unexecuted checks, and do not replace successful results with speculation. Prioritize first launch, normal use, boundary inputs, and quit/relaunch, and explain which limitations are unfinished work.
\end{framed}
\endgroup